%% file: PaperForReview.tex
\documentclass[10pt,twocolumn,letterpaper]{article}

\usepackage{wacv}              

\usepackage{graphicx}
\usepackage{amsmath}
\usepackage{amssymb}
\usepackage{booktabs}
\usepackage[accsupp]{axessibility}
\usepackage{algorithm,algpseudocode}

\usepackage[pagebackref,breaklinks,colorlinks]{hyperref}

\usepackage[capitalize]{cleveref}
\crefname{section}{Sec.}{Secs.}
\Crefname{section}{Section}{Sections}
\Crefname{table}{Table}{Tables}
\crefname{table}{Tab.}{Tabs.}

\def\wacvPaperID{1547} 
\def\confName{WACV}
\def\confYear{2024}

\newcommand{\eat}[1]{}

\begin{document}

\title{ReCLIP: Refine Contrastive Language Image\\Pre-Training with Source Free Domain Adaptation}

\author{
Xuefeng Hu$^{1}$
\and
Ke Zhang$^{2}$
\and
Lu Xia$^{2}$
\and
Albert Chen$^{2}$
\and
Jiajia Luo$^{2}$
\and
Yuyin Sun$^{2}$
\and
Ken Wang$^{2}$
\and
Nan Qiao$^{2}$
\and
Xiao Zeng$^{2}$
\and
Min Sun$^{2}$
\and
Cheng-Hao Kuo$^{2}$
\and
Ram Nevatia$^{1}$
\and 
{\small $^{1}$University of Southern California}
\space 
{\small $^{2}$Amazon}
\\
{\tt\small $^{1}$\{xuefengh,nevatia\}@usc.edu}
\\
{\tt\small $^{2}$\{kezha,luxial,aycchen,lujiajia,yuyinsun,zixiaow,qiaonan,zenxiao,minnsun,chkuo\}@amazon.com}
}

\maketitle

\begin{abstract}
   Large-scale pre-trained vision-language models (VLM) such as CLIP\cite{radford2021learning} have demonstrated noteworthy zero-shot classification capability, achieving 76.3\% top-1 accuracy on ImageNet without seeing any examples. However, while applying CLIP to a downstream target domain, the presence of visual and text domain gaps and cross-modality misalignment can greatly impact the model performance. To address such challenges, we propose ReCLIP, a novel source-free domain adaptation method for VLMs, which does not require any source data or target labeled data. 
   ReCLIP first learns a projection space to mitigate the misaligned visual-text embeddings and learns pseudo labels. Then, it deploys cross-modality self-training with the pseudo labels to update visual and text encoders, refine labels and reduce domain gaps and misalignment iteratively. 
   With extensive experiments, we show that ReCLIP outperforms all the baselines significantly and improves the average accuracy of CLIP from 69.83\% to 74.94\% on 22 image classification benchmarks. Code available at \url{https://github.com/michiganleon/ReCLIP_WACV}.
\end{abstract}
\vspace{-1em}

\section{Introduction}
    \input{sections/introduction}

\section{Related Works}
    \input{sections/background}

\section{Method}
    \label{sec:method}
    \input{sections/method}

\section{Experiment and Results}
    \label{sec:experiments}
    \input{sections/experiments}

\section{Conclusion}
    \input{sections/conclusion}

{\small
\bibliographystyle{ieee_fullname}
\bibliography{egbib}
}

\newpage \
\newpage

\input{sections/supp}

\end{document}

%% file: sections/introduction.tex
\label{sec:introduction}

Large-scale pre-training vision-language models (VLM) such as CLIP\cite{radford2021learning} have emerged recently and have formed a new paradigm in the task of image classification. Instead of annotating images with class labels, vision-language models match images towards text embeddings from their category names. With semantic relationship from text and large-scale pre-training over 400 million image-caption pairs, CLIP is capable of performing accurate image classification on novel target domains requiring zero training samples but only a dictionary of potential category names. 

\begin{figure}[t]
    \centering
    \includegraphics[width=\linewidth]{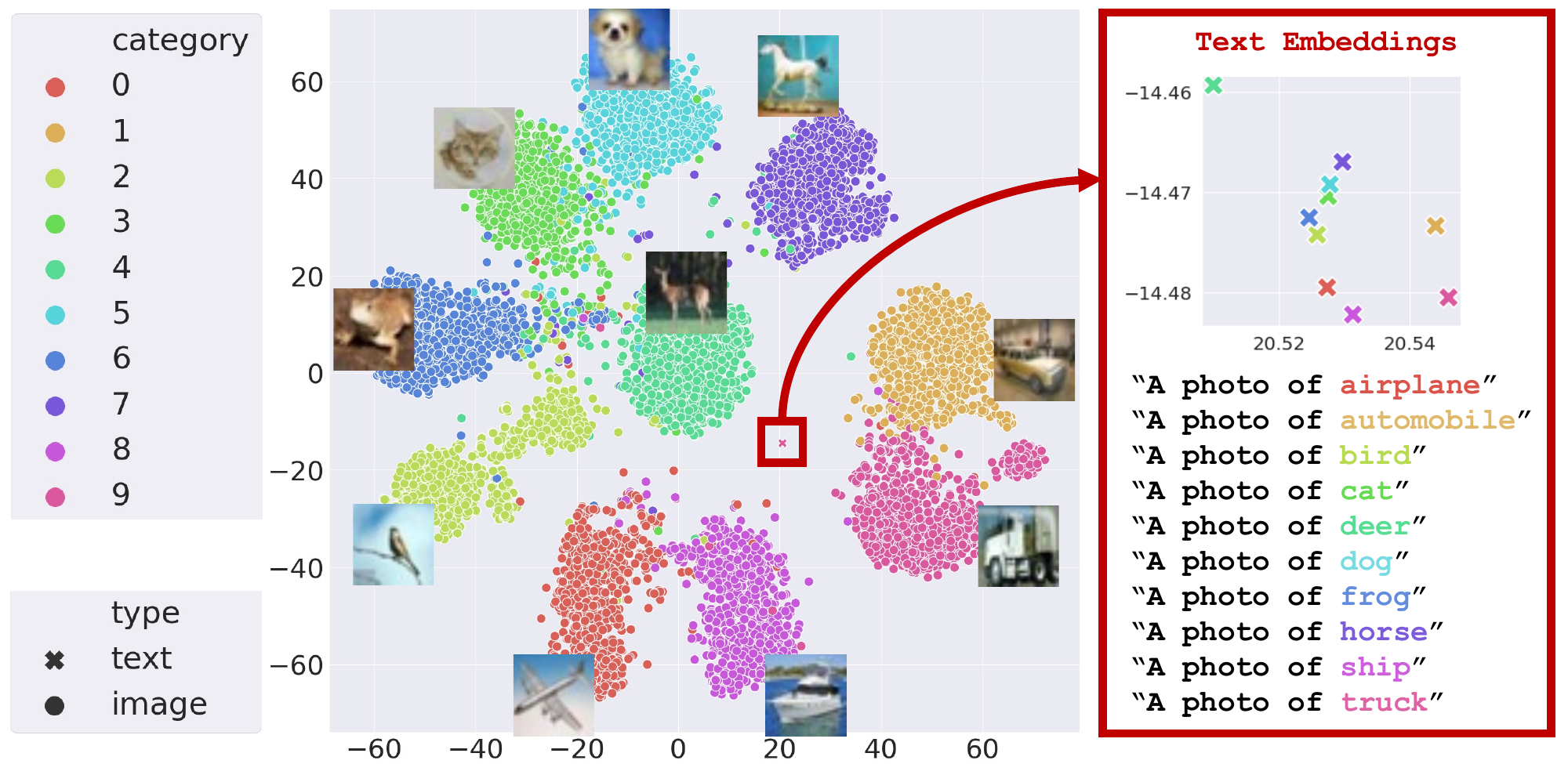}
    \caption{
    the t-SNE plot of visual and text embeddings from CLIP on CIFAR10~\cite{cifar} test set. It is clear to see the misalignment in the vision-language space: the text embedding of a class name is adjacent to ones of other classes, but distant from image embeddings in the same class.
    }
    \vspace{-1em}
    \label{fig:intro}
\end{figure}

However, we still observe domain gaps from both image and text input that impact CLIP performance. The existence of visual domain gap between source and target images has been a challenge for computer vision models \cite{wang2018deep,csurka2017domain}. \eat{Although CLIP has impressive generalization ability that benefited from the large-scale training,} CLIP has been observed to have limitations on visual embedding when data comes from less common domains, \eg PatchCamelyon\cite{pcam}, CLEVR\cite{johnson2017clevr}, \etc. 
On the other hand, the domain gap in text is also a challenge for vision-language models. The performance of CLIP is often limited by the text embeddings rather than the visual embeddings, especially on fine-grained datasets \eg RESISC45\cite{resisc45}, Birdsnap\cite{berg2014birdsnap}, where CLIP is able to create distinctive visual embeddings but the text embeddings from class names fail to capture discriminative information. 

In addition to the gaps in the visual and text domains, we have identified significant misalignment between visual and text embeddings across most datasets. Some recent studies \cite{liang2022mind,tanwisuth2023pouf} have also observed similar modality gaps across various contrastive-learned visual-language models.
Figure~\ref{fig:intro} provides examples of this issue on the widely used benchmark CIFAR10. We believe that there are two primary reasons for these misalignments. Firstly, text embeddings may be redundant, as CLIP was trained to work with millions of captions and concepts, whereas target domain categories might only activate limited feature dimensions, leaving the remaining ones inactive and redundant; these redundant dimensions can dominate the similarity calculation. Secondly, visual embeddings may contain a significant amount of class-agnostic information; since CLIP uses real captions for training, it preserves rich information, such as lighting, color, texture, and relationship, but only a small portion of this information is crucial for classification. 


Therefore, adaptation on both visual and text representations, and re-alignment between visual and text embeddings are crucial in improving the target domain performance of vision-language models like CLIP. 
However, traditional domain adaptation methods have significant limitations in this context. One major challenge is that these methods either require target domain labeled examples (e.g. semi-supervised domain adaptation\cite{yao2015semi,saito2019semi,daume2010frustratingly}), or source domain examples (e.g., unsupervised domain adaptation \cite{kang2019contrastive,na2021fixbi,sharma2021instance}). Nonetheless, typical use cases of CLIP only have access to unlabeled target images, which requires source-free unsupervised domain adaptation that does not need source data or labeled target data. 
Another challenge is that existing methods assume conditions that may not hold for vision-language models. For instance, most existing methods \cite{liang2020we,wang2020tent,yang2022attracting} assume a lightweight classifier, while a vision-language model uses a large text encoder to generate classification weights based on category descriptions. Such modules add flexibility and complexity to adaptation. 
Thus, the lack of labeled data from source and target domains and the presence of multiple adaptable modules make it essential to develop a novel source-free domain adaptation algorithm for vision-language models. 

More recently, POUF\cite{tanwisuth2023pouf} also proposes to address the misaligned embeddings of a vision-language model through source-free adaptation. However, the unsupervised objective of POUF considers each target example independently, instead of taking advantages from the neighboring relationship
over the entire embedding space. Moreover, POUF cannot leverage multiple template augmented text embeddings as used in CLIP and our proposed method, which limited its performance during the adaptation. 

To take advantage of the unified vision-language space, and address the challenges on the visual and text domain gaps and cross-modality misalignment, we propose ReCLIP, a novel source-free domain adaptation method to \textbf{Re}fine \textbf{CLIP} models. 
Firstly, ReCLIP addresses the misalignment of visual and text embeddings from CLIP by learning a projection subspace that removes redundant dimensions and class-agnostic information, and realigns embeddings. ReCLIP then utilizes the neighboring relationship between aligned embeddings, and employs label propagation to produce accurate pseudo-labels in the target domain. 
Secondly, ReCLIP leverages cross-modality self-training with high-confidence pseudo labels to iteratively refine embedding spaces and label assignments. Two parallel components are deployed to update the text and visual encoders. The first component fine-tunes the text encoder while freezing the visual to pull the text embedding of a label closer to the embeddings of images assigned the label. Meanwhile, the second component fine-tunes the visual encoder so that the images under the same label get closer to each other and to the multi-template augmented text embedding of the label. During fine-tuning, each component learns cross-modality consistency in the target domain, leading to new label assignments. ReCLIP selects labels agreed upon by both components as high-confidence ones for the next iteration. This iterative process improves the quality of visual and text embeddings and significantly enhances the assignment 
of pseudo labels.

Our contributions are summarized in the following:
\begin{itemize}
\itemsep -0.1em 
    \item We proposed ReCLIP, a novel source-free domain adaptation method for vision-language model, which enhances the CLIP’s classification ability towards target domains without labeled data;
    \item We identified the cross-modality misalignment issue between CLIP’s visual and language embeddings, and address the issue with an efficient projection-based component in ReCLIP;
    \item We proposed a novel cross-modality self-training algorithm with high quality commonly agreed pseudo labels leveraging cross-modality consistency to mitigate domain gaps from both visual and text inputs; 
    \item With extensive experiments and ablation studies, ReCLIP produces consistent and significant improvements over CLIP and other baseline methods; ReCLIP improves the average accuracy of CLIP from 69.83\% to 74.94\% on 22 datasets.
\end{itemize}


%% file: sections/background.tex
\subsection{Large-Scale Vision-Language Models}\label{limitation_clip}
    
    Many large-scale pre-training vision-language models have been recently proposed and demonstrate impressive zero-shot classification ability, such as CLIP~\cite{radford2021learning}, ALIGN~\cite{jia2021scaling} that perform large-scale contrastive training for strong generalization ability, and DeCLIP~\cite{csurka2017domain}, SLIP~\cite{mu2022slip} that focus on efficient training with additional self-supervised objectives. In this work, we adopt CLIP as our main base model, as it is still the most representative vision-language model with outstanding zero-shot classification performance and publicly available model weights. In addition, we will also demonstrate the effectiveness of our method with different base models in ablation studies. 


    \noindent\textbf{Augmented prompts through multiple templates. } 
    CLIP makes classification prediction by matching the visual embeddings of query images with the text embeddings of categories names (wrapped in template text such as ``\texttt{A photo of a \{\}}''), and selects the category with the highest cosine similarity as prediction (please refer to the supplementary materials for more details on CLIP and VLM).

    To further align these text embeddings with the pre-training distribution generated from real captions, CLIP prepares a long list of templates with various contexts for each of the 27 benchmarks it evaluated on. Instead of using just one template, CLIP reported scores are produced with the averaged text embeddings from a list of templated prompts for each category to boost performance. 
    
    \noindent\textbf{Limitations of CLIP.} We observe the following conditions where CLIP's performance could be improved. 
    \textbf{1) Inaccurate Text Description.} The accuracy of CLIP can sometimes be drastically improved when the classification weights are fully-supervised fine-tuned, e.g., On EuroSAT, accuracy of CLIP improved from 59.9\% to 98.2\%~\cite{radford2021learning}. This indicates that CLIP has good quality default visual representations, but the zero-shot performance is limited by the quality of text-generated classification weights. This is often observed on fine-grained datasets (e.g., AID~\cite{xia2017aid}, FGVC~\cite{fgvc}, EuroSAT~\cite{helber2019eurosat}, \etc), where the class names can not fully capture the visual differences between classes (e.g., ``737-200" and ``747-200" as class names from FGVC);
    \textbf{2) Visual Gap.} On some datasets, there are clear gaps for CLIP to be further improved even after the fully supervised fine-tuning on classification weight. For example, fine-tuned CLIP achieves only 42.9\% on Country211~\cite{radford2021learning}, and 85.97\% on PatchCamelyon\cite{pcam} (a binary classification task with state-of-the-art system achieves 97.50\%). This indicates that the visual encoder of CLIP can also be further improved. \textbf{3) Visual-Text Misalignment.} Recent studies \cite{liang2022mind, udandarao2022understanding} have also shown that the modality gap between visual and text embeddings caused by contrastive pretraining could also limit the performance of CLIP. By modifying contrastive temperature during pre-training~\cite{liang2022mind}, or by minimizing the gap during few-shot fine tuning~\cite{udandarao2022understanding}, these works suggest that mitigating the modality gap can benefit the classification ability of CLIP.

\subsection{Unsupervised Domain Adaptation}
\label{shot:intro}
Unsupervised Domain Adaptation (UDA) is a task aimed at improving target domain performance of models that were pre-trained on a related but different source domain. Many techniques have been developed~\cite{kang2019contrastive,na2021fixbi,sharma2021instance,sun2019unsupervised,wei2021metaalign}, including a recent method designed for visual-language models\cite{lai2023padclip}. However, most of these techniques are not ideal for the purpose of improving CLIP's zero-shot performance, as they often require access to source domain data, while we do not require access to CLIP's training data. 

\noindent\textbf{Source-Free Adaptation} defines a more challenging setting than UDA, where training examples are not available in both the source and target domains. SHOT~\cite{liang2020we} is one of the first Source-Free Adaptation (SFDA) methods. SHOT updates the feature extractor with cluster-based pseudo labels and information entropy loss, while maintaining the classifier frozen. AaD~\cite{yang2022attracting} improves SHOT by replacing the information entropy loss with a novel Attracting-and-Dispersing (AaD) loss. This simple but effective approach achieves state-of-the-art performance on the task of SFDA. 

More recently, POUF \cite{tanwisuth2023pouf} also proposes to mitigate the misalignment embeddings through source-free domain adaptation for vision-language models. But the optimization objective of POUF has limited its performance in two ways: 1) the training of POUF imposes dependency on the number of text encoder inputs (prompts), which limits POUF from using multiple templates to boost performance, especially on datasets with large number of classes; 2) the training objectives consider each image separately and fail to leverage neighboring relationships.



%% file: sections/method.tex
\begin{figure*}[ht]
        \centering
        \includegraphics[width=\textwidth]{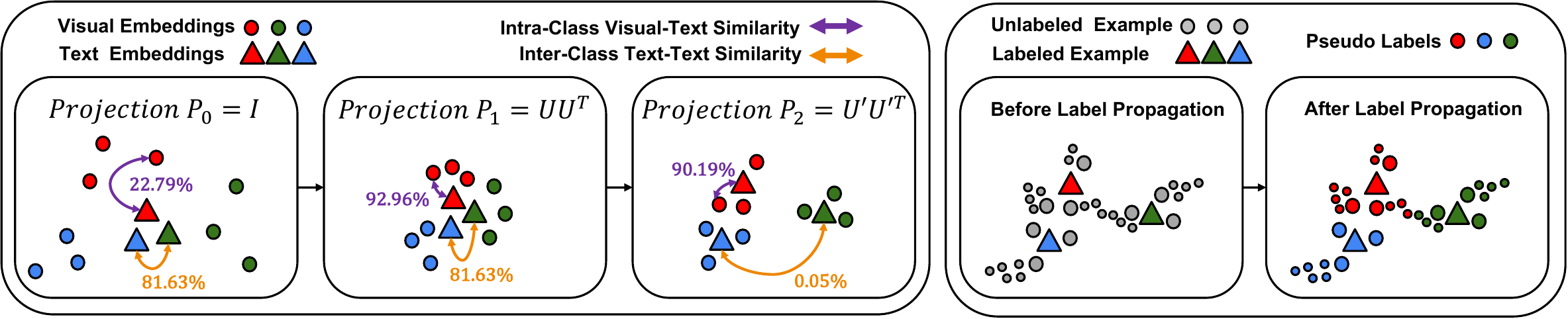}
        \caption{Demonstration on Feature Redundancy Removal (left) and Label Propagation (right). \textbf{Left}: $P_0$ shows the original distribution of visual and text embeddings of CLIP, where text embeddings are close to each other distant from visual embeddings; $P_1=UU^{\top}$ removes the class agnostic information from visual embeddings, and has pulled closer visual and text embeddings. $P_2=U'U'^\top$ separates the text embeddings away by removing the redundant information from them. Similarity values demonstrated in this example is calculated based on average statistics from CIFAR10 test set; \textbf{Right}: the Label Propagation process generates pseudo labels for unlabeled training images by propagating label information from labeled text embeddings (categories names) to unlabeled visual embeddings (training images) through nearest-neighbor connections.}
        \label{fig:project}
        \vspace{-1em}
    \end{figure*}

We describe our method \textbf{ReCLIP}, which \textbf{Re}fines \textbf{CLIP}'s classification performance by accessing only to the pre-trained model and the following target domain data:
\begin{itemize}
\itemsep 0em 
    \item Pre-trained vision-language model $M=\{M_T, M_V\}$, with text encoder $M_T$ and visual encoder $M_V$,
    \item Unlabeled target images $X=\{x_1, x_2,...,x_n\}$,
    \item Target class names $C=\{c_1,c_2,...,c_m\}$. 
\end{itemize}

Our goal is to increase the classification accuracy of $M$ on target data $X$. As the first method that studies the source-free adaptation problem for vision-language model, we approach this problem in two steps: (1) How to align visual and text embeddings by removing class-agnostic and redundant information in a learned projection space (Section ~\ref{sec:ps}). Then we show how to assign pseudo labels for images in the projection space via label propagation (Section~\ref{sec:pl}); (2) How to utilize the pseudo labels to further mitigate the visual and text domain gaps by efficiently updating both visual and text encoders, we propose a cross-modality self-training algorithm which updates embeddings and pseudo labels in a iterative fashion (Section~\ref{alg}).


\subsection{Projection Space to Align Visual and Text} 
\label{sec:ps}

Figure~\ref{fig:intro} demonstrates the misalignment issue of text and visual embeddings from CIFAR10~\cite{cifar}, which we have also observed over all the ablation datasets.  The plot indicates that the text embeddings of different class names are closer to each other than to images in the corresponding categories. We also validate the misalignment with quantitative statistics, as shown in Figure~\ref{fig:project}. The average cosine similarity between text embeddings is $82\%$, while the average similarity between visual and text embeddings from the same category is only $23\%$. This indicates that the unified vision-language space of CLIP is far from well aligned.

As highlighted in Section \ref{sec:introduction}, although the visual and text embeddings from CLIP convey rich information, much of them could be redundant and class-agnostic to target classification tasks. This redundancy can result in misalignment between the text and visual embeddings. We hence propose a projection-based method to eliminate the redundancy from both visual and text embeddings.

\eat{
A possible explanation is, CLIP representation space was trained to cooperate with millions of linguistic and visual concepts. 
On one hand, the category names from the same dataset might only activate a tiny portion of the feature dimensions that are relevant to this test domain, while the remaining dimensions are inactivated and redundant. In this case, text embeddings of class names might become very similar to each other as the inactivated dimensions dominate the similarity calculation. 
On the other hand, CLIP uses real caption text embedding to match with images during the training stage, and therefore it will tend to preserve information from many perspectives (e.g. color, texture, relationship, etc) on visual embeddings, while only a small portion of such information might be useful when performing classification on the specific test set. In this case, the similarity between visual and text embeddings might be low, as a large portion of visual embeddings could be class-agnostic information. 
}

\noindent\textbf{Remove class-agnostic information from visual embeddings.} A straightforward way to remove class-agnostic information from visual features is just to project all the visual embeddings onto the span of text embeddings. 
Assuming we have a $d$ dimensional representation space $\mathcal{R}^d$, and we have $m$ classes whose text embeddings are $T = [t_1,...,t_m]\in \mathcal{R}^{m\times d}$, where $t_i=M_t(c_i)$ for $i\in\{1,2,...,m\}$. With Singular Value Decomposition 
$$U, S, V = svd(T)$$
we get $U=[e_1, e_2, ..., e_m]$ as the orthonormal basis of the span of $T$, which defines a projection matrix $P_1=UU^\top$. Then, $\forall f\in\mathcal{R}^d$, we can calculate $f'=fP_1$ with
\begin{align*}
    e_k \cdot (f - f') &= 0, \forall k\in\{1,...m\}
\end{align*}
where $f - f'$ is the class-agnostic information that does not contribute to the classification. As shown in Figure \ref{fig:project}, $P_1$ increases the average similarity between images and text embeddings from the same category to 92.96\% on CIFAR10.

\noindent\textbf{Remove redundant information from text embeddings}. As suggested in Principal Component Analysis, the first dimension $e_1$ of the outer-space basis $U$ will be the major component that most $\{t_1,...,t_m\}$ overlap on. Removing the major component $e_1$ will make all text embeddings nearly perpendicular to each other. Therefore, with $U'=[e_2,e_3,...,e_m]$ we define a new projection matrix $P_2=U'U'^\top$. As shown in Figure \ref{fig:project}, $P_2$ successfully separates the text embeddings from different classes to an average cosine similarity of 0.05\%, while maintaining high intra-class visual-text similarity at 90.19\% on CIFAR10.  

In addition to the improvement of CIFAR10 statistics, experiments on pseudo label generation also indicate the effectiveness of embedding space induced by $P_2$ in improving clustering performance, as demonstrated in Section \ref{abla:cluster}.



    \begin{figure*}[ht]
        \centering
        \includegraphics[width=\textwidth]{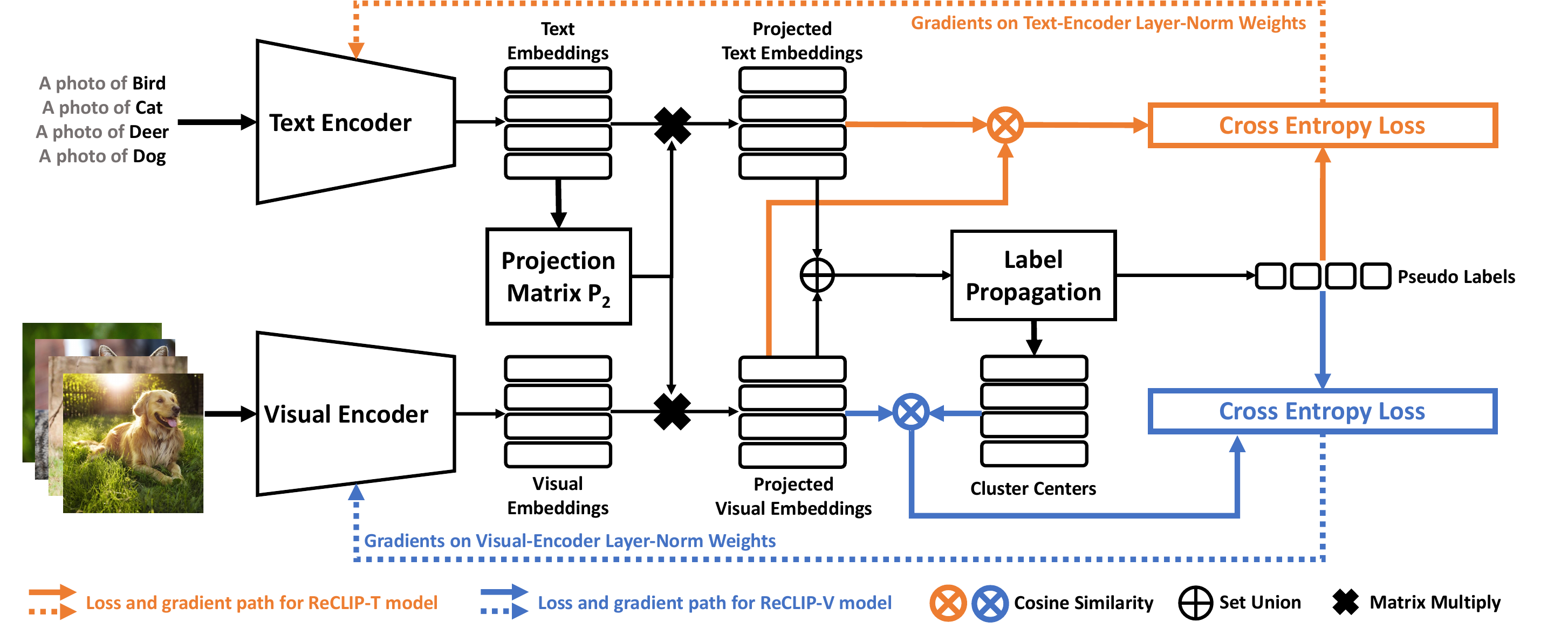}
        \caption{Flow Chart of ReCLIP-V and ReCLIP-T. Orange symbols describe the loss and gradients path of ReCLIP-V, and blue symbols describe the loss and gradients path of ReCLIP-T. Black symbols describe the common steps that both ReCLIP-V and ReCLIP-T have. }
        \label{fig:model1}
    \end{figure*}

        \begin{figure}[ht]
    \centering
    \includegraphics[width=\linewidth]{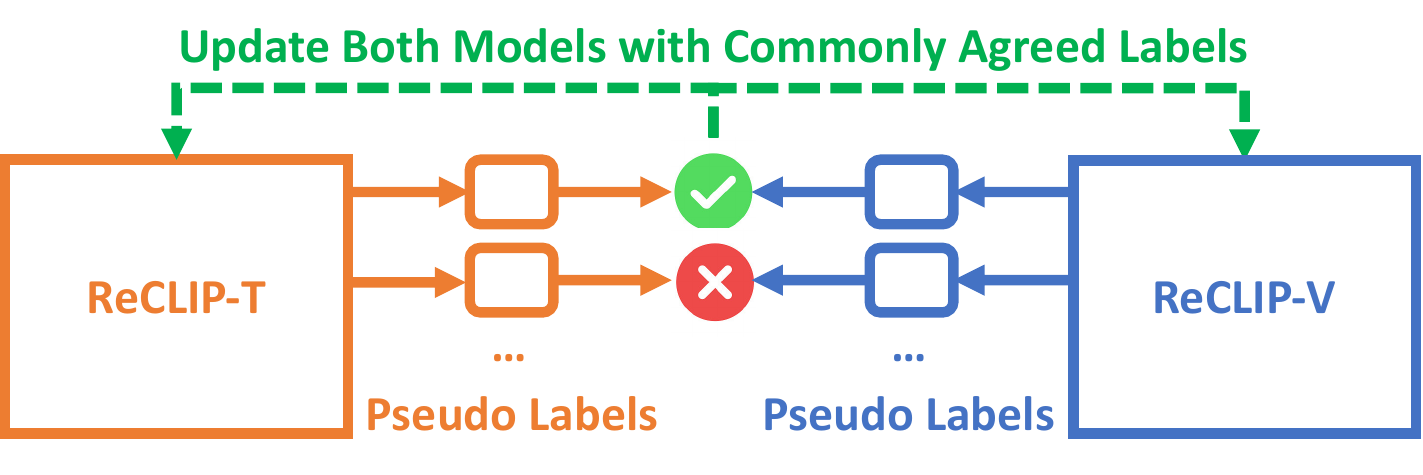}
    \caption{Flow Chart of Pseudo Labels Sharing. The cross-modality self-training algorithm merges the pseudo labels from ReCLIP-T and ReCLIP-V at the end of each epoch and updates the encoders only on high-confidence pseudo labels agreed by both.}
    \label{fig:model3}
    \vspace{-1em}
\end{figure}

\subsection{Pseudo Label Generation for VLM}
\label{sec:pl}
\eat{Label Propagation \cite{iscen2019label} is a semi-supervised learning method that generates pseudo label assignments by propagating label information from labeled to unlabeled data points through nearest neighbor connections. In our task, there is originally no access to labeled data points to perform such semi-supervised training. 

However, with the improved visual-text neighboring condition through the embedding alignment $P_2$, it is now possible for us to perform label propagation by treating text embeddings from class names as labeled points, and visual embeddings from images as unlabeled points.}

The projection matrix $P_2$ removes the redundancies and aligns visual and text embeddings\eat{ into a space with improved neighboring relationships}, which enables the generation of pseudo labels through Label Propagation \cite{iscen2019label}, which is a semi-supervised learning method that propagates label information from labeled to unlabeled data points through nearest neighbor connections, as demonstrated in Figure \ref{fig:project}.
Although in source-free adaptation we do not have access to labeled data points, the embedding alignment through $P_2$ has enabled us to treat text embeddings from class names as labeled points, and visual embeddings from images as unlabeled points.


With labeled examples $\{\hat{t}_i\}_{i=1}^m$ (class name embeddings) and unlabeled examples $\{\hat{v}_j\}_{j=1}^n$ ( image visual embeddings), we make the union set $L$:
$$L = [\hat{t}_1,\hat{t}_2,...,\hat{t}_m,\hat{v}_1,\hat{v}_2,...,\hat{v}_n]\in\mathcal{R}^{d\times(m+n)}$$
Following Label Propagation \cite{iscen2019label}, we first produce affinity matrix $A_k$ through $k-$nearest neighbor affinity ranking
$
    A_k = top^k(L^\top L)
$
where $top^k(\cdot)$ is an operation that keeps the top $k$ highest value per row from the full affinity matrix $L^\top L$. Then, with normalization and symmetrization, we have:
\begin{align*}
    \mathcal{W} &= D^{-\frac{1}{2}}(A_k + A_k^\top)D^{-\frac{1}{2}}
\end{align*}
where $D:=diag(W\mathbf{1}_{m+n})$ is the degree matrix, $\mathbf{1}_{m+n}$ is the all-ones $(m+n)-$vector, and $\mathcal{W}$ is the normalized adjacency matrix that defines the random walk probability. With an label matrix $Y\in\mathcal{R}^{(m+n)\times m}$ is defined with elements
$$
    Y_{ji} := 
    \begin{cases}
        1, & \text{if } j=i, j\leq m\\
        0, & \text{otherwise}
    \end{cases}
$$
where $Y_{ji}$ is 1 for the text embedding entries at the corresponding column, and 0 otherwise. 
Then, the pseudo label vector $Z$ can be estimated by solving the random walk problem with initial state $Y$, propagation probability matrix $\mathcal{W}$ and diffusion magnitude $\alpha$:
\begin{equation}
    Z := (\mathbf{I}-\alpha\mathcal{W})^{-1} Y \label{cg}
\end{equation}
where $(\mathbf{I}-\alpha\mathcal{W})^{-1}$ is the closed-form solution to the random walk problem. As $(\mathbf{I}-\alpha\mathcal{W})\in\mathcal{R}^{m+n}$ is not sparse, and therefore the calculation of its inverse matrix is very time consuming, we use conjugate gradient (CG) to approximately solve Equation \ref{cg}, following the suggestion from \cite{iscen2019label}.
Finally, with Equation \ref{cg} solved, the pseudo label can be given by
$$
\Tilde{y}_j := \arg \max_i z_{m+j,i}
$$
where $\Tilde{y}_j$ is the pseudo label of image $x_j$, and $z_{ji}$ is the $(j,i)$ element of matrix $Z$.

\subsection{Source-Free Adaptation for Vision-Language Model via Cross-Modality Self-Training}
\label{alg}
    \eat{Different from the adaptation methods on classic image classification models which adapt only one model, }
    Vision-language models present a new challenge to adaptation algorithms, where both visual and text encoders need to be adapted. In this section, we discuss how to mitigates the domain gaps of visual and text domains, and propose a cross-modality self-training algorithm with pseudo labels from~\ref{sec:pl} to iteratively update the label assignments, and the visual and text encoders.
    
    The self-training algorithm of ReCLIP consists of two parallel components: 
    ReCLIP-T aims at closing the text domain gap by pushing text embeddings towards visual embeddings of the same class, by fine-tuning the text encoder with the visual encoder frozen.
    ReCLIP-V aims at closing the visual domain gap by pushing visual embeddings of the same class closer to each other, by fine-tuning the visual encoder with the text encoder frozen.
    On top of ReCLIP-V and ReCLIP-T, we integrate the commonly-agreed pseudo labels to produce high-confidence training signals. For inference, we add the prediction logits from both ReCLIP-V and ReCLIP-T to make the final prediction.
        

\noindent\textbf{ReCLIP-T: Text Encoder Training.} We optimize the text encoder $M_t$ with simple cross-entropy loss $Loss^T:=CE(\hat{Y}^T,\Tilde{Y})$ between pseudo label $\Tilde{Y}$ and cosine similarity prediction logits $\hat{Y}^T=[\hat{v}_1,...,\hat{v}_n]^\top[\hat{t}_1,...,\hat{t}_m]$. 
The objective of adaptation on the text encoder is to push text embeddings $\{\hat{t}_i\}$ closer to the image embeddings $\{\hat{v}_j\}$ from the same class based on pseudo label assignments  $\Tilde{Y}^T$. In Figure \ref{fig:model1} we present the details of ReCLIP-T, the detailed algorithm is provided in the supplementary materials.

\noindent\textbf{ReCLIP-V: Visual Encoder Training.} 
The goal of visual encoder adaptation is to push visual embeddings $\{\hat{v}_j\}$ from the same class to be closer to each other, to form a better feature space for classification. As contrastive loss is expensive and applying constraints on batch size, we have instead chosen to push visual embeddings closer to the center of its class instead of other visual embeddings as an alternative resort. To be specific, in ReCLIP-V we optimize the visual encoder $M_v$ with cross-entropy loss $Loss^V:=CE(\hat{Y}^V,\Tilde{Y})$ between pseudo label $\Tilde{Y}$ and cosine similarity logits $\hat{Y}^V=[\hat{v}_1,...,\hat{v}_n]^\top[\hat{w}_1,...,\hat{w}_m]$, where $\hat{w}_1,...,\hat{w}_m$ are the class centers calculated based on $\Tilde{Y}$. In Figure \ref{fig:model1} we present the details of ReCLIP-V, the detailed algorithm is provided in the supplementary materials.

\noindent\textbf{High-Confidence Pseudo Labels Sharing.} ReCLIP-V updates the similarities among visual embeddings with $Loss^V$, while ReCLIP-T updates the projection matrix and text embeddings with $Loss^T$. As these two modules separately optimize the visual and text encoders with different objectives, their pseudo labels may start to diverge after a certain number of epochs, resulting in different views where only the commonly agreed samples are likely to be correctly classified. As such, ReCLIP collects pseudo labels from both ReCLIP-V and ReCLIP-T at the end of each epoch, and updates both models with only the commonly agreed pseudo labels $\Tilde{Y}$, as illustrated in Figure~\ref{fig:model3}. The detailed algorithm is provided in the supplementary materials.

\eat{\noindent\textbf{High-Confidence Pseudo Labels Sharing.} In ReCLIP-V, the location and neighboring relationships between visual embeddings will be updated by $Loss^V$. In ReCLIP-T, the projection matrix and location of text embeddings will be updated by $Loss^T$. With ReCLIP-V and ReCLIP-T separately updating CLIP's visual and text encoders with different optimization goals, their pseudo labels might start to diverge after certain epochs, and creates two different views where the commonly agreed samples might have a higher chance to be correct. Thus in ReCLIP we collect pseudo labels from both ReCLIP-V and ReCLIP-T at the end of each epoch, and update both models with only commonly agreed pseudo labels $\Tilde{Y}$, as shown in Figure~\ref{fig:model3}. The detailed algorithm is also provided in the supplementary materials.}


%% file: sections/experiments.tex
\begin{table*}[ht]
\small
\resizebox{\textwidth}{!}{
\setlength{\tabcolsep}{0.2em} 
{\renewcommand{\arraystretch}{1.2}
\begin{tabular}{ccccccccccccccccccccccccc}
              & \rotatebox{90}{Avg Acc} & \rotatebox{90}{AID\cite{xia2017aid}} & \rotatebox{90}{Birdsnap\cite{berg2014birdsnap}} & \rotatebox{90}{Caltech101\cite{caltech101}} & \rotatebox{90}{CIFAR10\cite{cifar}} & \rotatebox{90}{CIFAR100\cite{cifar}} & \rotatebox{90}{Country211\cite{radford2021learning}} & \rotatebox{90}{DTD\cite{dtd}} & \rotatebox{90}{EuroSAT\cite{helber2019eurosat}} & \rotatebox{90}{FER2013\cite{fer}} & \rotatebox{90}{FGVC\cite{fgvc}} & \rotatebox{90}{Flowers\cite{flower102}} & \rotatebox{90}{Food101\cite{bossard2014food}} & \rotatebox{90}{GTSRB\cite{GTSRB}} & \rotatebox{90}{ImageNet\cite{deng2009imagenet}} & \rotatebox{90}{MNIST\cite{deng2012mnist}} & \rotatebox{90}{Oxford Pet\cite{pet}} & \rotatebox{90}{PCam\cite{pcam}} & \rotatebox{90}{SST2\cite{radford2021learning}} & \rotatebox{90}{RESISC45\cite{resisc45}} & \rotatebox{90}{Cars\cite{cars}} & \rotatebox{90}{STL10\cite{stl}} & \rotatebox{90}{SUN397\cite{xiao2010sun}}  \\ \hline
CLIP-\textit{report}    & 70.08 & -     & 48.30 & 92.6* & 96.20 & 77.90 & \textbf{32.70}    & 55.30 & 59.90 & 57.50 & 36.1* & 78.7* & 92.90 & 50.30 & 75.30 & 87.20 & 93.50 & 58.80 & 64.00 & 71.60 & 77.3* & 99.30 & 67.70 \\
CLIP-\textit{single}    & 65.53 & 61.30 & 51.88 & 92.02 & 95.19 & 77.18 & 25.78             & 52.50 & 56.03 & 52.22 & 30.18 & 74.19 & 92.56 & 45.57 & 73.46 & 52.63 & 93.21 & 57.75 & 52.39 & 63.29 & 76.45 & 99.47 & 66.42 \\
CLIP-\textit{multi}     & 69.83 & 68.73 & 52.48 & 91.63 & 95.60 & 78.22 & 31.84             & 55.37 & 60.00 & 56.39 & 31.59 & 79.04 & 93.08 & 50.59 & 75.52 & 76.23 & 93.62 & 62.43 & 68.92 & 69.66 & 77.88 & 99.36 & 67.97 \\ \hline
AaD                     & 46.53 & 69.83 & 52.42 & 91.45 & 96.54 & 80.18 & 0.47              & 55.43 & 11.12 & 16.91 & 32.37 & 78.61 & 0.99  & 51.26 & 0.11  & 89.81 & 93.62 & 49.95 & 49.92 & 2.51  & 0.52  & 99.41 & 0.25  \\
AaD \textit{peak}       & 71.79 & 70.33 & 52.58 & 91.93 & 96.55 & 80.46 & 31.90             & 55.59 & 76.18 & 55.67 & 32.43 & 79.22 & 93.04 & 52.83 & 75.53 & 91.95 & 93.73 & 64.03 & 68.97 & 71.01 & 77.96 & 99.42 & 67.96 \\ \hline
POUF                    & 69.73 & 64.83 & 52.91 & 92.97 & 96.06 & 80.39 & 28.19             & 56.65 & 67.95 & 55.92 & 32.88 & 75.62 & 92.71 & 51.47 & 73.05 & 91.22 & 94.20 & 66.57 & 48.22 & 67.54 & 76.72 & 99.50 & 68.38 \\
POUF \textit{peak}      & 69.76 & 64.87 & 52.96 & 92.97 & 96.06 & 80.39 & 28.22             & 56.75 & 67.95 & 55.92 & 32.91 & 75.62 & 92.73 & 51.47 & 73.06 & 91.22 & 94.20 & 66.75 & 48.60 & 67.54 & 76.72 & 99.53 & 68.38 \\ \hline
ReCLIP                  & 74.94 & 77.97 & 52.96 & 93.02 & 96.95 & 82.32 & 31.92             & 60.85 & 78.75 & 58.07 & 36.63 & 82.05 & 94.15 & 66.81 & 75.81 & 90.88 & 95.61 & 70.15 & 73.48 & 78.41 & 77.96 & 99.58 & 74.41 \\ 
ReCLIP \textit{peak}    & \textbf{75.85} & \textbf{79.27} & \textbf{53.28} & \textbf{93.10} & \textbf{97.04} & \textbf{83.42} & 31.95 & \textbf{61.38} & \textbf{79.94} & \textbf{58.29} & \textbf{38.70} & \textbf{83.14} & \textbf{94.18} & \textbf{69.14} & \textbf{76.01} & \textbf{97.11} & \textbf{96.05} & \textbf{70.56} & \textbf{73.48} & \textbf{79.31} & \textbf{79.26} & \textbf{99.59} & \textbf{74.53} \\ \hline
\end{tabular} 
}
}
\caption{Classification accuracies (\%) on 22 benchmarks. * on FGVC, Caltech101, Oxford-IIIT Pet and Flowers102, CLIP reported mean-class-accuracy. All other scores in this table are top-1 accuracy.}
\vspace{-1.5em}
\label{tab:main}
\end{table*}

\begin{table}
\resizebox{\linewidth}{!}{
\begin{tabular}{cccccc}
              & Avg       & Ar            & Cl        & Pr        & Rw     \\ \hline
CLIP \textit{single}                     & 82.45          & 82.70          & 68.10          & 89.10          & 89.90          \\ \hline
POUF-prompt              & 84.28          & 83.70          & 71.20          & 91.40          & 90.80          \\
POUF                & 86.10          & \textbf{86.20} & 73.80          & 92.70          & 91.70          \\ \hline
Label Propagation & 84.94          & 83.27          & 73.49          & 91.89          & 91.09          \\
ReCLIP  & \textbf{87.00} & 86.11          & \textbf{75.97} & \textbf{93.90} & \textbf{92.01} \\ \hline
\end{tabular}
}
\caption{Comparison of ReCLIP and published scores from POUF\cite{tanwisuth2023pouf} on Office-Home\cite{venkateswara2017deep}, both use CLIP-\textit{single} as base model.}
\label{office}
\vspace{-0.6em}
\end{table}

\noindent\textbf{Baselines} We use the following methods for comparison:

\noindent\textbf{1) CLIP \cite{radford2021learning}:} State-of-the-art zero-shot image classification model. We choose CLIP with ViT/L-14 architecture as the main baseline model for comparison and adaptation. 
We report both published results from Radford \etal ~\cite{radford2021learning} and our reproduction, denoted as \textit{report} and \textit{multi} respectively. Both \textit{report} and \textit{multi} are prepared with the official prompt template lists provided by Radford \etal ~\cite{radford2021learning}. In addition, we also report the results we reproduced with a single template (``\texttt{A photo of a \{\}}''), denoted as \textit{single};
    
\noindent\textbf{2) AaD \cite{yang2022attracting}:} State-of-the-art SFDA method. We adapt the official code to apply it on CLIP and our benchmarks;

\noindent\textbf{3) POUF\cite{tanwisuth2023pouf}:} A recent SFDA method that also aims to mitigate misaligned visual and text embedding spaces. Since POUF does not report on the benchmarks where CLIP has published scores, we produce its results on these benchmarks using its official code. We report the best performing version of POUF which fine-tunes the entire model.

\noindent\textbf{Evaluation and Datasets.} 
\textbf{1) Main Results:} for SFDA comparison between ReCLIP, POUF, AaD and base model CLIP, we use an abundant and comprehensive list of 21 common image classification benchmarks out of the 27 benchmarks from Radford \etal~\cite{radford2021learning}, except the 6 datasets where CLIP are evaluated on the custom splits or protocols which are not released at the time of this submission (KITTI~\cite{kitti}, UCF101~\cite{soomro2012ucf101}, VOC2007~\cite{pascal-voc-2007}, Kinetics700~\cite{kinetics}, HatefulMemes~\cite{kiela2020hateful}, CLEVR~\cite{johnson2017clevr}). In addition to the ablation dataset AID~\cite{xia2017aid} we use for hyper-parameters selection, SFDA evaluation is performed on 22 benchmarks in total. 
\textbf{2) Comparison with POUF:} For additional comparison with POUF on its published scores, we evaluate ReCLIP on Office-Home \cite{venkateswara2017deep}, which contains four different domains: Art (Ar), Clipart (Cl), Product (Pr) and Real-World (Rw). 
\textbf{3) Ablation Studies:} we choose AID, CIFAR10, CIFAR100 and SUN397 as ablation datasets to represent datasets with different sizes and characteristics. 
For more details on evaluation datasets, please refer to supplementary materials. 

For SFDA evaluation in Section~\ref{sec:main}, AaD and ReCLIP use CLIP-\textit{multi} as base model, and POUF uses CLIP-\textit{single} due to its design. For experiments on Office-Home, both ReCLIP and POUF use CLIP-\textit{single} as base model. 

Unless otherwise specified, we perform our experiments in transductive manner, where SFDA methods ReCLIP, POUF and AaD first perform adaptation on the unlabeled test data of each dataset, and then the adapted models are evaluated on the same test data following the standard CLIP inference protocol. For all benchmarks, we use top-1 classification accuracy as our metric,



\noindent\textbf{Implementation Details}
\label{details}
For the self-training of ReCLIP, we fine-tune the layer-normalization \cite{ba2016layer} weights with other weights frozen, as it is shown to be one of the most effective and stable option to adapt models with noisy supervision~\cite{wang2020tent}. 
For the SFDA evaluation, we use AID~\cite{xia2017aid} to select the best hyper-parameter for ReCLIP, POUF and AaD. We then use the same set of hyper-parameters for all 22 datasets during the evaluation. We match the maximum adaptation steps for all methods to be the same, as $\min\{5000 \text{ iterations}, 50 \text{ epochs}\}$. 
For the evaluation on Office-Home, we select the hyper-parameter on the Real-World (Rw) domain and use the same hyper-parameters across all domains for evaluation. 
For details on exact hyper-parameters used in experiments, ablation studies on choices of learnable modules, and the setup of Label Propagation, please refer to supplementary materials.

\subsection{Main Results}
\label{sec:main}

In Table \ref{tab:main} we present the SFDA accuracy of ReCLIP, AaD and POUF over 22 datasets. Besides the accuracy from the final epoch of self-training, we report the accuracy from the peak-performing epoch for AaD, POUF and ReCLIP as well, denoted as \textit{peak}.

ReCLIP achieves consistent and significant improvements over CLIP on 21 datasets and comparable performance on Country 211. ReCLIP improves the averaged top-1 accuracy of CLIP by 5.11\% and 6.02\% at the \textit{final} and \textit{peak} epochs respectively over the 22 datasets without accessing any labeled data, which outperforms both baseline adaptation methods AaD, POUF by clear margin. 


AaD achieves 1.96\% improvements over CLIP at its \textit{peak} epochs. However, it encounters drastic performance drops at \textit{final} epochs that lose 25.26\% of the averaged accuracy, due to collapsed unsupervised training on target datasets such as Food101, SUN397, ImageNet, etc. Meanwhile, ReCLIP maintains the performance at \textit{final} epochs, with only 0.91\% differences from the \textit{peak} epochs. These results suggest the effectiveness of the high-quality commonly agreed pseudo labels of ReCLIP in stabilizing the noisy self-training and preventing model collapse.


POUF achieves 4.20\% improvement over its base model CLIP-\textit{single}. However, such improvement is counteracted by the inability to employ multiple prompts to enhance text embedding quality, as suggested by CLIP~\cite{radford2021learning}. Multiple templates create a large number of prompts, which are not likely to fit in the same mini-batch for text encoder optimization. ReCLIP also experiences this limitation when fine-tuning the text encoder. However, thanks to the dual-component structure of ReCLIP, although ReCLIP-T also only use single template for text-encoder optimization, ReCLIP-V can still take advantage of the multiple template augmented text embeddings and provides better pseudo labels to ReCLIP-T through pseudo-label sharing. 
In addition to the advantage brought by multi-template augmented text embeddings, ReCLIP also takes advantage from the neighboring relationships over the entire visual-text embedding space, while POUF does not, which has also contributed to the better performance of ReCLIP. More evidence and discussion on this are covered in Section~\ref{sec:dahome}.

Country211 is designed to predict geo-location based on visual appearance, while CLIP might tend to describe the image from actual content and texture. As shown in \cite{radford2021learning}, CLIP can only achieve 42.9\% after its classifier is fine-tuned in the fully supervised way. Therefore, it is challenging to obtain improvement during source-free domain adaptation. 

\begin{table}[t]
    \resizebox{\linewidth}{!}{
    \begin{tabular}{ccccc}
    \multicolumn{1}{l}{}            & CIFAR10        & CIFAR100       & AID             & SUN397    \\ \hline
    Vanilla CLIP                    & 95.54          & 76.48          & 64.87           & 67.25     \\
    Label Propagation               & 96.38          & 80.66          & 74.73           & 70.54     \\
    ReCLIP-V                        & 96.69          & 80.84          & 79.47           & 67.15     \\
    ReCLIP-T                        & 96.50          & 81.10          & 79.07           & 70.12     \\
    ReCLIP (w/o Label Sharing)      & 97.40          & 82.80          & 80.01           & 71.10     \\
    ReCLIP (w/ Label Sharing)       & \textbf{97.48} & \textbf{84.14} & \textbf{82.53}  & \textbf{71.34} \\ \hline  
    \end{tabular}
    }
    \caption{Comparison of classification accuracy with different version ReCLIP on ablation datasets. ReCLIP with Label Sharing (Figure \ref{fig:model3}) is shown to be most effective compared to ReCLIP-V, ReCLIP-T (Figure \ref{fig:model1}) and their simply assembled predictions (ReCLIP w/o Label Sharing).}
    \label{abla:method}
    \vspace{-0.6em}

\end{table}

\subsection{Comparison with POUF}
\label{sec:dahome}
In Table \ref{office} we present the comparison between the published scores of POUF and ReCLIP on the Office-Home, where both methods use CLIP-\textit{single} (ViT/B-16) as base model. We also include the Label Propagation pseudo label accuracy generated on our projected CLIP embeddings prior to any updates on the base model. It is shown that the Label Propagation accuracy already outperforms POUF-prompt, which fine-tunes the learnable text prompt. Moreover, ReCLIP achieves clear improvement over POUF over most of the domains, with 2.17\%$\uparrow$ on Cl, 1.20\%$\uparrow$ on Pr, 0.31\%$\uparrow$ on Rw and on-par performance on Ar. These results indicate that ReCLIP can still outperform POUF without using multi-template augmented embeddings.

\begin{table}[t]
    \resizebox{\linewidth}{!}{
    \begin{tabular}{lcccc}
                                & AID   & CIFAR10   & CIFAR100 & SUN397  \\ \hline
    Vanilla CLIP                & 68.80                     & 95.59                         & 78.21              & 67.97           \\ \hline
    Hierarchical Clustering     & 55.20                     & 36.52                         & 9.27               & 46.93           \\
    Spectrum Clustering         & 68.10                     & 61.25                         & 57.35              & 27.45           \\
    $k$-means Clustering        & 72.73                     & 95.07                         & 49.43              & 43.66           \\ \hline
    $k$-NN Classifier ($P_0$)     & 72.30                     & 93.74                         & 69.46            & 60.72             \\
    $k$-NN Classifier ($P_1$)     & 72.76                     & 95.77                         & 77.81            & 63.07             \\
    $k$-NN Classifier ($P_2$)     & 72.43                     & 95.76                         & 78.19            & 63.29             \\ \hline
    Label Propagation ($P_0$)   & 60.80                     & 94.01                         & 63.58              & 51.77           \\ 
    Label Propagation ($P_1$)   & 60.43                     & 96.23                         & 45.41              & 33.41          \\ 
    Label Propagation ($P_2$)   & \textbf{76.36}            & \textbf{96.31}                & \textbf{81.56}     & \textbf{70.44}           \\ \hline
    \end{tabular}
    }
    \caption{Pseudo label accuracy with different methods. Label Propagation on projection space $P_2$ is shown to be the most effective and stable method in generating accurate pseudo labels. }
    \label{table:clustering}
    \vspace{-0.6em}
    \end{table} 

\subsection{Ablations Studies}
\label{sec:abla}
    In this section, we present the ablation studies on comparison of different ReCLIP versions, pseudo label generation, and ablation on various VLMs as base models. 
    We use AID, CIFAR10, CIFAR100 and SUN397 as our ablation datasets, and the test set of each dataset is equally split  into two fixed partitions. We report the ablation results in an inductive manner where models are first adapted on partition 1 and then evaluated on partition 2. Note that results in this section are not directly comparable to \ref{sec:main} because the different evaluation partition.
    
    \subsubsection{Effectiveness of ReCLIP Components}
    In Table \ref{abla:method} we present the comparison between different versions of ReCLIP. As shown, Label Propagation can create pseudo labels with significantly improved accuracy compared to vanilla CLIP. On the top of Label Propagation, both ReCLIP-V and ReCLIP-T (Figure~\ref{fig:model1}) are shown to be effective in providing further improvements. 
    In ReCLIP(w/o Label Sharing) we present the result by simply assembling predictions from separately trained ReCLIP-V and ReCLIP-T at inference time. Comparing the last two rows of Table \ref{abla:method} we observe that ReCLIP~(w/ Label Sharing) has clear improvement over ReCLIP~(w/o Label Sharing), which indicates that the commonly agreed pseudo-labels stabilizes the noisy adaptation process and improved both ReCLIP-V and ReCLIP-T to achieve better performance.

    \subsubsection{Comparison on Pseudo Label Generations}\label{abla:cluster}
    
    In Table \ref{table:clustering}, we compare methods in pseudo label generation.
    For clustering based methods, we assign the same pseudo labels for examples from the same cluster, based on the in-cluster majority vote; 
    For $k$-NN Classifier and Label Propagation methods, we experiment them on original CLIP feature space $P_0$, and on projection spaces $P_1, P_2$ as described in Figure \ref{fig:project}.
    For $k$-NN Classifiers, we assign each example with the major vote prediction within its $k$-nearest-neighborhood, with $k$ equal to the average sample count per class.  
    For Label Propagation on $P_0$, we select the example with the highest confidence from each class as the labeled example to perform label propagation as a baseline. 
    Label Propagation on $P_1, P_2$ are as described in Section \ref{sec:ps}.

    Table \ref{table:clustering} indicates $k$-NN based methods achieve better performance on projection spaces $P_1$ are $P_2$, which indicates the effectiveness of $P_1, P_2$ in refining CLIP's visual embeddings. On Label Propagation methods, $P_2$ gives a significant improvement over $P_0, P_1$, indicating its effectiveness in aligning CLIP's visual and text embeddings.

    \begin{table}[t]
    \setlength{\tabcolsep}{0.2em}
    \resizebox{\linewidth}{!}{
    \begin{tabular}{ccccc}
                      & CIFAR10                                     & CIFAR100                                      & AID                                   & SUN397     \\
                      & Init  $\rightarrow$   Adapt            & Init $\rightarrow$ Adapt                 & Init $\rightarrow$ Adapt         & Init         $\rightarrow$ Adapt           \\ \hline
    SLIP (ViT-L/16)   & 89.45     $\rightarrow$   \textbf{91.80}    & 56.69  $\rightarrow$ \textbf{67.61}           & 48.13   $\rightarrow$\textbf{64.07}   & 55.56   $\rightarrow$ \textbf{65.28} \\
    DeCLIP (ViT-B/32) & 90.57     $\rightarrow$   \textbf{94.50}    & 66.58  $\rightarrow$ \textbf{77.10}           & 53.53  $\rightarrow$\textbf{65.93}    & 63.05   $\rightarrow$ \textbf{66.90} \\ \hline
    CLIP (RN50)       & 71.46     $\rightarrow$   \textbf{82.73}    & 42.32  $\rightarrow$ \textbf{53.15}           & 53.43  $\rightarrow$\textbf{65.97}    & 59.76   $\rightarrow$ \textbf{65.38}      \\
        CLIP (ViT-B/32)   & 89.83     $\rightarrow$   \textbf{92.15}    & 65.25  $\rightarrow$ \textbf{71.09}       & 60.83  $\rightarrow$\textbf{76.80}    & 62.96   $\rightarrow$ \textbf{68.30} \\ \hline
    \end{tabular}
    }
    \caption{Ablation Studies on the effectiveness of ReCLIP on different model architecture and pre-training strategies.}
    \label{effectiveness}
    \vspace{-0.5em}
    \end{table}

    \subsubsection{Comparison on other Vision-Language Models}

    ReCLIP is designed to improve the classification performance of visual-language models in general, not only on CLIP. We tested the effectiveness of ReCLIP on SLIP\cite{mu2022slip} and DeCLIP\cite{li2021supervision}, both of these improved CLIP by adding self-supervision learning objectives during pre-training. We have also tested ReCLIP on other versions of CLIP with smaller architectures. As shown in Table \ref{effectiveness}, ReCLIP demonstrates steady and significant improvements on various vision-language models and architectures.

    \subsubsection{Runtime and Inductive Performance} 
    Self-training of ReCLIP is very efficient, which completes adaptation in only 0.5 to 5 GPU-Hour on a single V100 GPU, depending on the target dataset size. Note that this adaptation time is a one-time effort on each target domain and ReCLIP can then inference on unseen data from the same domain without re-training. For complete runtime of ReCLIP over each benchmarks and more inductive evaluation results, please refer to the supplementary materials. 

%% file: sections/conclusion.tex
In this paper, we introduce ReCLIP, a novel solution on source-free domain adaptation for vision-language models.  
ReCLIP first uses a novel designed projection space to re-aligns visual and text embeddings and to generate dependable pseudo labels for target classification tasks. ReCLIP further applies cross-modality self-training with pseudo labels, which iteratively enhances label assignments and visual and text embeddings. Compared to the previous methods AaD and POUF, ReCLIP provides an effective and stable solution to the source-free adaptation problem of vision-language models. ReCLIP significantly improves CLIP, increasing the average accuracy from $69.83\%$ to $74.94\%$ across 22 datasets.

%% file: sections/supp.tex


\section*{Appendix I: Background on CLIP}
    
\begin{figure}[h]
        \centering
        \includegraphics[width=\linewidth]{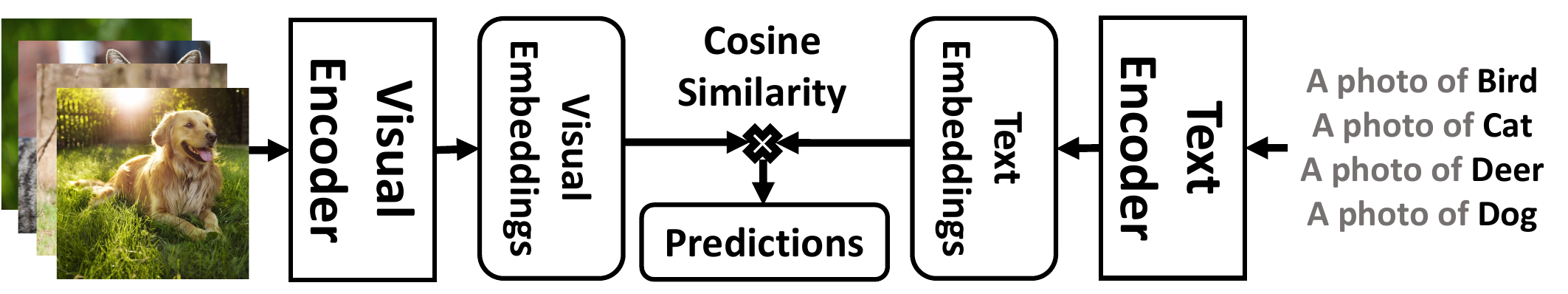}
        \caption{CLIP performs classification on target classes by comparing visual embeddings with the text embeddings generated from class names.}
        \label{fig:clip}
    \end{figure}

    CLIP performs contrastive learning over 400 millions web-retrieved pairs of images and captions by pulling the visual and text representation near if they are from the same pair and away if they are not. At inference stage, CLIP makes classification prediction by matching the visual embeddings of query images with the text embeddings of categories names (wrapped in template text such as ``\texttt{a photo of \{\}}'', or a list of templates and uses the averaged embedding, as discussed in the main paper), and selects the category with the highest cosine similarity as prediction, as shown in Figure \ref{fig:clip}. CLIP is capable of performing classification over novel tasks without any training example, as long as the category names are provided. CLIP has demonstrated outstanding zero-shot classification accuracy, e.g. 76.3\% top-1 accuracy on ImageNet without seeing any examples from the dataset. \cite{radford2021learning}.
    
\section*{Appendix II: Algorithms}

    As described in Section 3.3 of the main paper, ReCLIP is composed of two parallel components that are designed for visual and text encoder fine-tuning, namely ReCLIP-V and ReCLIP-T. On top of ReCLIP-T and ReCLIP-V, we integrate the pseudo labels by filtering the commonly-agreed ones to produce high-confidence training signals for both sides. In this Section, we present the detailed description of ReCLIP-T and ReCLIP-V in Algorithm \ref{alg:1}, and the pseudo label sharing in Algorithm \ref{alg:3}.

    \begin{algorithm*}[t]
    \begin{algorithmic}
    \Require Vision Language Pre-trained Model $M=\{M_v,M_t\}$
    \Require Unlabeled Images $X=\{x_1,...,x_n\}$
    \Require Class Names $C=\{c_1,...,c_m\}$
    \Require Mode = ReCLIP-V or ReCLIP-T
    \Comment{ReCLIP-V updates $M_v$ with $M_t$ frozen}\\
    \Comment{ReCLIP-T updates $M_t$ with $M_v$ frozen}
    \For{epoch $\gets 1$ to Max Epoch}
        \State $\{t_1,...,t_m\} \gets M_t(\{c_1,...,c_m\})$
        \State $\{v_1,...,v_n\} \gets M_v(\{x_1,...,x_n\})$
        \Comment{Calculate Visual and Text Embeddings}
        \State $U,S,V \gets svd([t_1,...,t_m])$, where $U=[e_1,...,e_m]$
        \State $P_2 \gets [e_2,...,e_m][e_2,...,e_m]^\top$ \Comment{Prepare Projection Matrix with Singular Value Decomposition}
        \State $\hat{t_i} \gets \frac{t_iP_2}{\lVert t_iP_2 \rVert}$
        \State $\hat{v_j} \gets \frac{v_jP_2}{\lVert v_jP_2 \rVert}$ 
        \Comment{Align Visual and Text Embeddings in Projection Space}
        \State $L \gets \{\hat{t_1},...,\hat{t_m},\hat{v_1},...,\hat{v_n}\}$
        \State $\Tilde{Y} \gets$ Label\_Propagation$(L)$  \Comment{Generate Pseudo Label through Label Propagation}
        \If{Mode=ReCLIP-T}
            \State $\hat{Y}\gets$ $[\hat{v_1},...,\hat{v_n}]^\top[\hat{t_1},...,\hat{t_m}]$ \Comment{Generate Predictions through Cosine-Similarity}
            \State Loss$^T$ $\gets$ Cross-Entropy$(\hat{Y}, \Tilde{Y})$
            \State Back-Propagation over $M_t$
        \ElsIf{Mode=ReCLIP-V}
            \State $w_i\gets \left(\sum_{\Tilde{Y}_j = i}v_j\right) / \left(\sum_{\Tilde{Y}_j = i} 1\right)$, for $i\in\{1,2,...,m\}$
            \State $\hat{w}_i \gets \frac{w_i}{\lVert w_i \rVert}$ for $i\in\{1,2,...,m\}$\Comment{Calculate the average embeddings for each class $i$}
            \State $\hat{Y}\gets$ $[\hat{v_1},...,\hat{v_n}]^\top[\hat{w_1},...,\hat{w_m}]$ \Comment{Generate Predictions through Cosine-Similarity}
            \State Loss$^V$ $\gets$ Cross-Entropy$(\hat{Y}, \Tilde{Y})$
            \State Back-Propagation over $M_v$
        \EndIf
    \EndFor
    \end{algorithmic}
    \caption{Visual and Text Encoder Self-Training: ReCLIP-V and ReCLIP-T}
    \label{alg:1}
    \end{algorithm*}
    
    \begin{algorithm*}[ht]
    \begin{algorithmic}
    \Require Component 1 $M^1=\{M_v^1,M_t^1\}$ (for ReCLIP-V), 
    \Require Component 2 $M^2=\{M_v^2,M_t^2\}$ (for ReCLIP-T)
    \Require Unlabeled Images $X=\{x_1,...,x_n\}$
    \Require Class Names $C=\{c_1,...,c_m\}$
    \State Self-Training Adaptation Stage:
    \For{epoch $\gets 1$ to Max Epoch}
        \State $\hat{Y}^1, \Tilde{Y}^1 \gets $ ReCLIP-V$(M^1, X, C)$
        \State $\hat{Y}^2, \Tilde{Y}^2 \gets $ ReCLIP-T$(M^2, X, C)$
        \Comment{ReCLIP-V/T generate predictions $\hat{Y}^{1},\hat{Y}^{2}$ and pseudo labels $\Tilde{Y}^{1},\Tilde{Y}^{2}$.}
        \State Commonly Agreed Index Map $Q \gets (\Tilde{Y}_1 = \Tilde{Y}_2)$
        \Comment{Boolean Index with $True$ indicates $\Tilde{Y}^1$ agrees with $\Tilde{Y}^2$.}
        \State Loss$^V$ $\gets$ Cross-Entropy$(\hat{Y}^1[Q], \Tilde{Y}^1[Q])$
        \State Loss$^T$ $\gets$ Cross-Entropy$(\hat{Y}^2[Q], \Tilde{Y}^2[Q])$
        \Comment{Only calculate loss on entries where $Q$ is True ($\Tilde{Y}^1$ agrees with $\Tilde{Y}^2$).}
        \State Back-Propagate $M^1_v$ with Loss$^V$
        \State Back-Propagate $M^2_t$ with Loss$^T$
    \EndFor
    \State
    \State Inference Stage: 
    \State $\hat{Y}^1 \gets $ ReCLIP-V$(M^1, X, C)$ \Comment{Generate inference predictions from ReCLIP-T/V}
    \State $\hat{Y}^2 \gets $ ReCLIP-T$(M^2, X, C)$ \Comment{At inference time, ReCLIP-T/V skip the pseudo label generation.}
    \State $\hat{Y} \gets \frac{1}{2}(\hat{Y}^1+\hat{Y}^2)$  \Comment{Aggregate prediction logits from both ReCLIP-T/V for prediction.}
    \State return $\arg\max\limits_{i} \hat{y}_{ji}$ as prediction for image $x_j$ \Comment{$Y=\{\hat{y}_{ji}\}$, where $\hat{y}_{ji}$ is probability of image $x_j$ on class $i$.}
    
    \end{algorithmic}
    \caption{ReCLIP with Pseudo Label Sharing}
    \label{alg:3}
    \end{algorithm*}

\section*{Appendix III: Evaluation Benchmarks}
    For the main result from the paper, we have evaluated our model as well as the baseline methods on the validation or test splits from 22 image classification benchmarks, according to the setup as stated from Radford, et al \cite{radford2021learning}.
    The 22 benchmarks is composed of the one ablation datasets AID~\cite{xia2017aid} that we used for hyper-parameter selection, and the 21 benchmarks (Caltech101\cite{caltech101},
CIFAR10\cite{cifar},
CIFAR100\cite{cifar},
ImageNet\cite{deng2009imagenet},
SUN397\cite{xiao2010sun},
Birdsnap\cite{berg2014birdsnap},
Country211\cite{radford2021learning},
DTD\cite{dtd},
EuroSAT\cite{helber2019eurosat},
FER2013\cite{fer},
FGVC\cite{fgvc},
Flowers\cite{flower102},
Food101\cite{bossard2014food},
GTSRB\cite{GTSRB},
MNIST\cite{deng2012mnist},
Oxford Pet\cite{pet},
PCam\cite{pcam},
SST2\cite{radford2021learning},
RESISC45\cite{resisc45},
Cars\cite{cars},
STL10\cite{stl}) from the 27 benchmarks CLIP reported in Radford, et al \cite{radford2021learning}, except: i) KITTI~\cite{kitti}, UCF101~\cite{soomro2012ucf101}, VOC2007~\cite{pascal-voc-2007}, Kinetics700~\cite{kinetics} that are object detection or video classification benchmarks that are out of the scope of our discussion; ii) HatefulMemes~\cite{kiela2020hateful} and CLEVR~\cite{johnson2017clevr}, where CLIP uses custom splits that are not released at the time of this submission. 
The detailed statistics on the number of images and the number of classes are reported in Table \ref{tab:meta}.

For comparison with POUF published score, we reported our scores on the Office-Home datasets. Office-Home contains 65 categories and 15588 images from four different domains: 2427 Art images, 4365 Clipart images, 4439 Product images and 4357 Real-World Images. 


    \begin{table*}[t]
    \small
        \resizebox{\textwidth}{!}{
        \setlength{\tabcolsep}{0.2em} 
        {\renewcommand{\arraystretch}{1.2}
            \begin{tabular}{cccccccccccccccccccccccccc}
                           & \rotatebox{90}{Average} & \rotatebox{90}{AID\cite{xia2017aid}} & \rotatebox{90}{Birdsnap\cite{berg2014birdsnap}} & \rotatebox{90}{Caltech101\cite{caltech101}} & \rotatebox{90}{CIFAR10\cite{cifar}} & \rotatebox{90}{CIFAR100\cite{cifar}} & \rotatebox{90}{Country211\cite{radford2021learning}} & \rotatebox{90}{DTD\cite{dtd}} & \rotatebox{90}{EuroSAT\cite{helber2019eurosat}} & \rotatebox{90}{FER2013\cite{fer}} & \rotatebox{90}{FGVC\cite{fgvc}} & \rotatebox{90}{Flowers\cite{flower102}} & \rotatebox{90}{Food101\cite{bossard2014food}} & \rotatebox{90}{GTSRB\cite{GTSRB}} & \rotatebox{90}{ImageNet\cite{deng2009imagenet}} & \rotatebox{90}{MNIST\cite{deng2012mnist}} & \rotatebox{90}{Oxford Pet\cite{pet}} & \rotatebox{90}{PCam\cite{pcam}} & \rotatebox{90}{SST2\cite{radford2021learning}} & \rotatebox{90}{RESISC45\cite{resisc45}} & \rotatebox{90}{Stanford Cars\cite{cars}} & \rotatebox{90}{STL10\cite{stl}} & \rotatebox{90}{SUN397\cite{xiao2010sun}}  \\ \hline
            Image Number   &   & 1500          & 2,149          & 9,146          & 10,000         & 10,000         & 21,100         & 1,880          & 5000          & 3,574          & 3,333          & 6,149          & 25,250         & 12,630         & 50,000         & 10,000         & 3,669          & 32,768         & 1,821          & 25,200         & 8,041          & 8,000          & 19,850         \\
            Class Number   &   & 30             & 500            & 102            & 10             & 100            & 211            & 47             & 10             & 8              & 100            & 102            & 102            & 43             & 1,000          & 10             & 37             & 2              & 2              & 45             & 196            & 10             & 397            \\ \hline
            AaD (h)    & 1.19 & 0.49 & 0.56 & 0.98 & 1.26 & 1.26 & 1.30 & 0.42 & 4.39 & 0.71 & 0.71 & 1.24 & 1.24 & 1.29 & 1.29  & 1.27 & 0.77 & 1.31 & 0.38 & 1.34 & 1.26 & 1.30 & 1.32 \\
            POUF (h)   & 6.18 & 4.51 & 7.07 & 5.61 & 5.80 & 5.71 & 7.30 & 5.50 & 5.60 & 3.73 & 5.02 & 5.82 & 6.38 & 6.41 & 13.58 & 5.74 & 4.13 & 6.79 & 4.91 & 6.33 & 5.97 & 5.92 & 8.19 \\
            ReCLIP (h) & 2.35 & 0.68 & 0.97 & 2.94 & 1.62 & 2.68 & 1.58 & 1.08 & 1.82 & 0.90 & 1.24 & 2.73 & 5.66 & 3.82 & 3.23  & 2.19 & 0.95 & 2.99 & 0.61 & 3.12 & 4.17 & 2.18 & 4.63 \\ \hline
            \end{tabular} 
        }
        }
    \caption{Metadata and Runtime comparison of AaD, POUF and ReCLIP of the 22 Evaluation Benchmarks. Time reported in the unit of hour (h).}
    \label{tab:meta}
    \end{table*}  

\section*{Appendix IV: Implementation Details}
    As mentioned in the main paper, we use AID to choose the best hyper-parameters for each baselines and evaluate them with the same hyper-parameters across the 22 datasets for SFDA evaluation. 
    
    For ReCLIP, we use learning rate of $10^{-3}$, weight decay of $10^{-4}$, momentum of $0.9$, batch size of 64, maximum length of $\min\{5000 \text{ iterations}, 50 \text{ epochs}\}$ and SGD optimization on both visual and text encoders. For Birdsnap, Country211, SUN397 and ImageNet which have more than 200 classes, we use a batch size of 32 due to large memory occupation from text inputs to fit the training on a single V100 GPU. For Label Propagation, we use propagation strength $\alpha=0.99$ and neighbor size $k=20$. 
    For datasets with more than 500 classes (Birdsnap, ImageNet), we notice the accuracy of pseudo labels generated by label propagation becomes unstable, and it requires additional hyper-parameter tuning to achieve good performance. To maintain stable performance, we turn off label propagation and simply use model predictions as pseudo labels on datasets with over 500 categories (Birdsnap, ImageNet). For all other datasets, we follow the exact process as described in Algorithm \ref{alg:1} and \ref{alg:3}.
    
    For both AaD and POUF, we have tested different hyper-parameters and report the the best performing setting, with learning rate of $10^{-3}$, weight decay of $10^{-3}$, momentum of $0.9$, SGD optimization on AaD, and learning rate of $10^{-2}$, weight decay of $10^{-3}$, momentum of $0.9$, SGD optimization on POUF. For both AaD and POUF, we extended their default training length to match our training length of ReCLIP, with batch size of 64 $\times$ $\min\{5000 \text{ iterations}, 50 \text{ epochs}\}$ steps on AaD, and batch size of 32 $\times$ $\min\{10000 \text{ iterations}, 100 \text{ epochs}\}$ steps on POUF.

    For ReCLIP on Office-Home, we use the Real-World (Rw) domain to choose the hyper-parameter. We use SGD optimizer with learning rate of $10^{-2}$ on the visual encoder and $10^{-3}$ on the text encoder, batch size of 64 and 5000 iteration as maximum step across all domains. For label propagation, we use $k=10$ due to the smaller dataset size. 

\section*{Appendix V: Additional Ablation Results}
    \subsection*{Choice on Learnable Modules}
        In Table \ref{modules}, we evaluate different learnable modules by comparing their fully-supervised fine-tuned performance. As suggested in \cite{wang2020tent}, fine-tuning the normalization weights is shown to be efficient and stable, compared to fine-tuning the entire weights in self-training of ReCLIP. 
    
        Recent research \cite{jia2022visual} as well as POUF~\cite{tanwisuth2023pouf} also suggests that learnable prompts can also be effective in providing stable and fast performance improvement during the fine-tuning of Transformer \cite{vaswani2017attention, dosovitskiy2020image} based models. In Table \ref{modules}, we perform Visual Prompt tuning following \cite{jia2022visual}, and our own designed Text Prompt. Please refer to Appendix VII for more details.

        As shown in Table \ref{modules}, fine-tuning Layer-Norm weights from Visual Encoder has the best fully supervised accuracy on both CIFAR10 and CIFAR100, while fine-tuning Layer-Norm weights from Text Encoder has the best fully supervised accuracy on AID. As described in Section 2 from the Main Paper, on some datasets (including AID), the performance of CLIP is mainly limited by the poor quality text embeddings from inaccurate class names. In this case, fine-tuning the text encoder will achieve better performance as we observed. Table \ref{modules} results suggest the necessity of fine-tuning CLIP from both the visual and text side to handle different scenarios. 

        \begin{table}[t]
        \resizebox{\linewidth}{!}{
            \begin{tabular}{ccccc}
                                                                & CIFAR10       & CIFAR100          & AID        & SUN397   \\ \hline
            Vanilla CLIP                                        & 95.54         & 76.48             & 64.87      & 67.25   \\
            Learnable Text Prompts                              & 97.50          & 82.18            & 93.73      & 75.27   \\
            Learnable Visual Prompts \cite{jia2022visual}       & 96.70          & 80.68            & 74.27      & 68.09   \\
            Text Encoder Layer-Norm                             & 97.32         & 83.30             & \textbf{94.8} & \textbf{78.47} \\
            Visual Encoder Layer-Norm                           & \textbf{97.8} & \textbf{85.16}    & 69.40        & 68.30 \\ \hline  
            \end{tabular}
        }
        \caption{Fully supervised fine-tuning accuracy of CLIP with different learnable modules on ablation datasets. On AID, fine-tuning weights from Text Encoder Layer-Norm is shown to be most effective; On CIFAR10 and CIFAR100, fine-tuning weights from Visual Encoder Layer-Norm is shown to be most effective.}
        \label{modules}
        \end{table}
    
    \subsection*{Inductive Results}

        \begin{table}
        \resizebox{\linewidth}{!}{
        \begin{tabular}{ccccc}
                              & CIFAR10 & CIFAR100 & AID   & SUN397 \\ \hline
        CLIP                  & 95.60   & 78.22    & 68.73 & 67.97  \\
        ReCLIP (Transductive) & 97.04   & 83.42    & 79.27 & 71.25  \\
        ReCLIP (Inductive)    & 96.92   & 82.30    & 79.87 & 74.53  \\ \hline
        \end{tabular}
        }
        \caption{Inductive and Transductive performance comparison of ReCLIP on ablation datasets. }
        \label{inductive}
        \end{table}

        We perform the SFDA evaluation in Table 1 from the main paper, to follow the protocols of AaD~\cite{yang2022attracting} and POUF~\cite{tanwisuth2023pouf} and to fully utilize the test examples. However, ReCLIP can also be applied in the inductive manner, so that the adaptation only has to be performed once for the target domain, and the adapted model will be effective on new and unseen examples of the target domain. In Table~\ref{inductive} we run ReCLIP in an inductive setting, where ReCLIP performs self-training on the training split of a dataset (0.5 to 5 GPU-Hour), and inference on the test split (similar to CLIP inference time). ReCLIP achieves similar improvements in the inductive setting as in the transductive setting. 
        
    \subsection*{Pseudo Label Quality}
        In Table \ref{tab:pseudolabel} we report the pseudo label accuracy of ReCLIP. We report the pseudo label accuracy from ReCLIP on the first epoch, before the self-training algorithm updates the model weights. 
        From Table \ref{tab:pseudolabel} we observe that the label propagation over projected visual and text embeddings has obtained ReCLIP pseudo labels with consistent improved accuracy over CLIP, only except Birdsnap and ImageNet which have more than 500 categories, as we discussed in Appendix IV. The results from Table \ref{tab:pseudolabel} demonstrate the effectiveness of our version of the label propagation method in generating reliable pseudo labels for vision-language models. More discussion on pseudo label generation is also covered in Section 4.3.2 of the main paper. 
        \begin{table*}[t]
        \small
        \resizebox{\textwidth}{!}{
        \setlength{\tabcolsep}{0.2em} 
        {\renewcommand{\arraystretch}{1.2}
        \begin{tabular}{cccccccccccccccccccccccccc}
                       & \rotatebox{90}{Average} & \rotatebox{90}{AID\cite{xia2017aid}} & \rotatebox{90}{Birdsnap\cite{berg2014birdsnap}} & \rotatebox{90}{Caltech101\cite{caltech101}} & \rotatebox{90}{CIFAR10\cite{cifar}} & \rotatebox{90}{CIFAR100\cite{cifar}} & \rotatebox{90}{Country211\cite{radford2021learning}} & \rotatebox{90}{DTD\cite{dtd}} & \rotatebox{90}{EuroSAT\cite{helber2019eurosat}} & \rotatebox{90}{FER2013\cite{fer}} & \rotatebox{90}{FGVC\cite{fgvc}} & \rotatebox{90}{Flowers\cite{flower102}} & \rotatebox{90}{Food101\cite{bossard2014food}} & \rotatebox{90}{GTSRB\cite{GTSRB}} & \rotatebox{90}{ImageNet\cite{deng2009imagenet}} & \rotatebox{90}{MNIST\cite{deng2012mnist}} & \rotatebox{90}{Oxford Pet\cite{pet}} & \rotatebox{90}{PCam\cite{pcam}} & \rotatebox{90}{SST2\cite{radford2021learning}} & \rotatebox{90}{RESISC45\cite{resisc45}} & \rotatebox{90}{Stanford Cars\cite{cars}} & \rotatebox{90}{STL10\cite{stl}} & \rotatebox{90}{SUN397\cite{xiao2010sun}}  \\ \hline
        CLIP \textit{repro} &     69.83   & 68.73          & 52.48          & 91.63          & 95.60          & 78.22          & 31.84          & 55.37          & 60.00          & 56.39          & 31.59          & 79.04          & 93.08          & 50.59          & 75.52          & 76.23          & 93.62          & 62.43          & 68.92          & 69.66          & 77.88          & 99.36          & 67.97          \\ \hline
        ReCLIP (pseudo label) &     72.54   & 74.50          & 43.25          & 91.91          & 96.56          & 81.40          & 26.30          & 59.04          & 73.36          & 57.15          & 36.33          & 82.55          & 93.95          & 60.64          & 25.11          & 82.85          & 94.77          & 62.46          & 68.86          & 77.63          & 77.66          & 99.52          & 70.54          \\ \hline
        
        \end{tabular} 
        }
        }
        \caption{ReCLIP pseudo label Quality. Results are generated with vanilla CLIP ViT-L/16 checkpoint, on the first epoch of ReCLIP before the training algorithms update the model weights.}
        \label{tab:pseudolabel}
        \end{table*}

\section*{Appendix VI: Time Analysis}
    We present the runtime required by SFDA methods, namely AaD, POUF and ReCLIP, in Table~\ref{tab:meta}. We matched all methods to be at the same training steps for fair comparison. As shown by the result, AaD takes an average of 1.19 hours to adapt, ReCLIP takes 2.35 hours and POUF takes 6.18 hours. ReCLIP is not much slower than AaD although ReCLIP trains two sets of encoders at the same time, except on datasets with more categories due to the time required for the Label Propagation process. However, POUF is much slower than both AaD and ReCLIP, due to its less efficient implementation. However, all three algorithms are very efficient as the adaptation only has to be applied once for each new target domain.

\section*{Appendix VII: Details on the design of learnable Language Prompt}

    \subsection*{What is Language Prompts}
       
       During the large-scale contrastive pre-training, CLIP~\cite{radford2021learning} was trained to match visual-text embedding between training images with their caption sentences such as \texttt{``A Golden Retriever dog sitting on grass''}. However, during inference time, category descriptions are usually provided in the form of phrases such as \texttt{``Golden Retriever''} or just  \texttt{``Dog''} instead of captions in complete sentences. To mitigate this gap, CLIP has proposed to use templates to wrap the category description phrase into complete sentences to generate better text embeddings. 

        For optimal performance, CLIP~\cite{radford2021learning} further claims that specific templates which provide contexts to the category names might help generate better text embeddings for classification. For example, CLIP finds the template prompt \texttt{``A photo of \{category name\}, a type of pet''} works the best for OxfordIII-Pet~\cite{pet}. CLIP has designed different lists of template prompts for all datasets it was evaluated on. The details can be found on their official GitHub repository \url{https://github.com/openai/CLIP/blob/main/data/prompts.md}.

    \subsection*{Learnable Language Prompts}
        
        As demonstrated by CLIP\cite{radford2021learning}, the wisely chosen template prompts might play a vital role in generating accurate text embeddings. However, this process largely depends on the heuristic design. Our goal for the learnable language prompt design is to make the prompt learnable and to avoid having different template prompts for different datasets. Additionally, this can also be an efficient and stable way to fine-tune the performance of CLIP. 

\begin{figure}
            \centering
            \includegraphics[width=\linewidth]{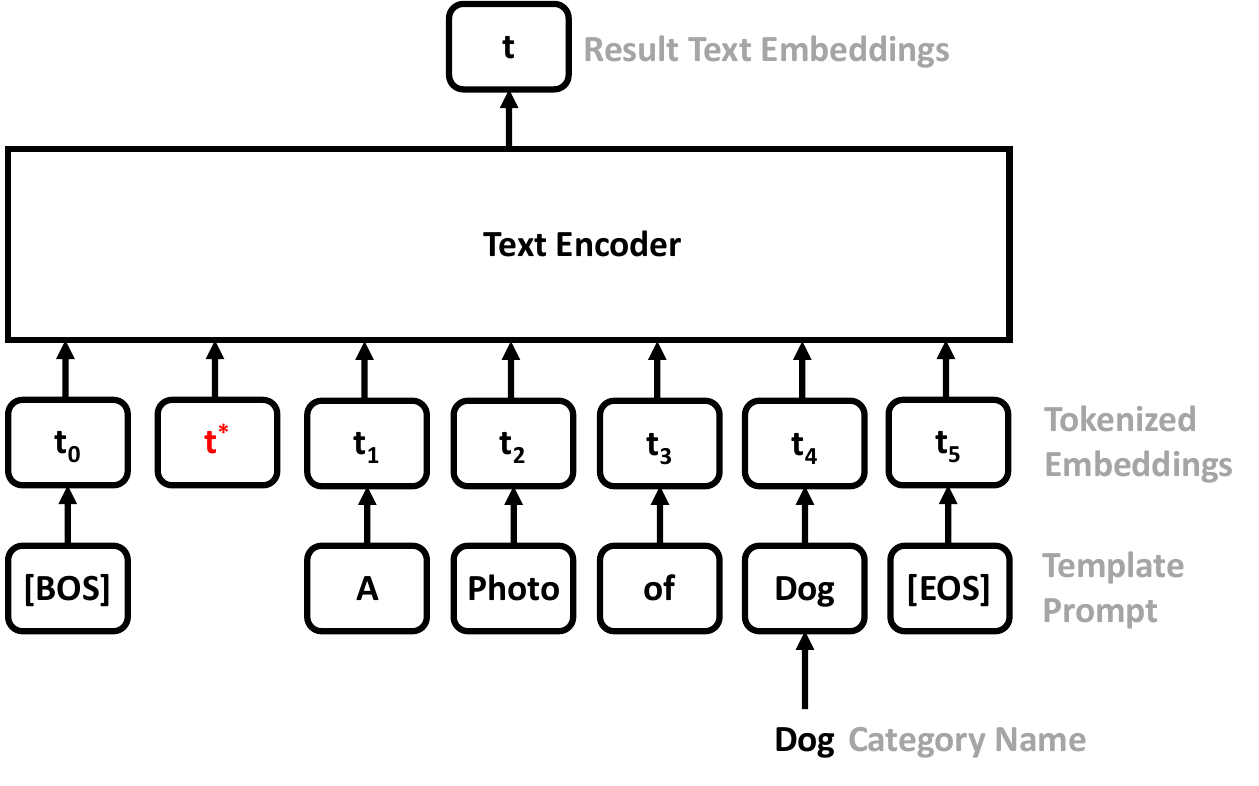}
            \caption{Demonstration of the design of Learnable Prompt. $\textcolor{red}{t^*}$ represents a learnable token embedding that is inserted at the beginning of the sequence of inputs to the transformer-based text encoder.  ``BOS'' and ``EOS'' stands for ``beginning of sentence'' and ``end of sentence'' and they serve as the special tokens for the text encoder to identify the beginning and end of the input sentence.}
            \label{fig:learnable}
        \end{figure}

        We start from the default template prompt \texttt{``A photo of \{category name\}''}, and insert an additional learnable token embedding $\textcolor{red}{t^*}$ at the beginning of the sentence, right after the Begin-Of-Sentence (BOS) token, as shown in Figure \ref{fig:learnable}. $\textcolor{red}{t^*}$ is initialized with the same embedding value of word \texttt{``is''} for reasonable performance before it is fine-tuned. During the fine-tuning process, token $\textcolor{red}{t^*}$ is made to be learnable while token embeddings for all other words are fixed. 

\section*{Appendix VIII: Limitation and Future Work}
    As mentioned in the Implementation Details section, we have observed that on datasets with more than 500 classes (Birdsnap, ImageNet), the accuracy of pseudo labels generated by label propagation becomes unstable, and it requires additional hyperparameter tuning to achieve good performance. To maintain stable performance, we have turned off label propagation and simply used model predictions as our pseudo labels on datasets with over 500 categories. Studies on how the hyper-parameters influence the label propagation performance on datasets with more than 500 categories will be important future work to further improve ReCLIP.

    Another future direction will be the utilization of augmentation consistency. Augmentation Consistency has been shown to be a very powerful unsupervised training signal and has been widely applied in unsupervised methods \cite{chen2020simple,chen2021exploring,he2020momentum}. Due to the scope and complexity of this project, we have not explored the usage of augmentation consistency in source-free domain adaptation. It will be important future work to explore the combination of the current ReCLIP with augmentation consistency to further improve the adaptation performance.